\pdfoutput=1
\documentclass[11pt]{article}

\usepackage[]{emnlp2021}

\usepackage{times}
\usepackage{latexsym}

\usepackage[T1]{fontenc}
\usepackage{newtxmath}

\usepackage[utf8]{inputenc}

\usepackage{microtype}

\usepackage{graphicx}  %Required
\usepackage{amsmath}
\usepackage{amsfonts}
\usepackage{paralist}
\usepackage{bbold}
\usepackage{bm}
\usepackage[toc,page]{appendix} 
\usepackage{float}
\usepackage{makecell}
\usepackage{subfig}

\DeclareSymbolFontAlphabet{\amsmathbb}{AMSb}%
\newcommand{\R}[0]{\amsmathbb R}

\newcommand{\E}[0]{\amsmathbb E}
\newcommand{\HE}[0]{\mathbf{H}^{(\text{enc})}}
\newcommand{\HD}[0]{\mathbf{H}^{(\text{dec})}}
\DeclareMathOperator*{\argmax}{\arg\!\max}
\DeclareMathOperator*{\layernorm}{LN}
\DeclareMathOperator*{\mha}{MHA}
\DeclareMathOperator*{\ffn}{FFN}
\DeclareMathOperator*{\relu}{ReLU}

\newcommand{\indep}{\perp \!\!\! \perp}

\usepackage{multirow}
\usepackage{paralist}
\usepackage{arydshln}
\usepackage{bibentry}
\usepackage{comment}

\makeatletter
\newcommand{\thickhline}{%
    \noalign {\ifnum 0=`}\fi \hrule height 1pt
    \futurelet \reserved@a \@xhline
}
\makeatother

\usepackage{color}
\usepackage{xcolor}
\definecolor{gblue}{RGB}{66,133,244}
\definecolor{gred}{RGB}{219,68,55}
\definecolor{gyellow}{RGB}{244,160,0}
\definecolor{ggreen}{RGB}{15,157,88}

\usepackage{algorithm}
\usepackage{algpseudocode}
\usepackage{bold-extra}
\usepackage{wrapfig}
\definecolor{deepcarminepink}{rgb}{0.94, 0.19, 0.22}
\definecolor{azure}{rgb}{0.0, 0.5, 1.0}

\algrenewcommand\alglinenumber[1]{\tiny #1:}

\newcommand\yumo[1]{\textcolor{red}{\textbf{#1}}}

\usepackage{mathtools}

\usepackage{todonotes}
\title{Text Summarization with Latent Queries}

\author{Yumo Xu \and Mirella Lapata\\
 Institute for Language, Cognition and Computation\\
 School of Informatics, University of Edinburgh\\
 10 Crichton Street, Edinburgh EH8 9AB\\
 \texttt{yumo.xu@ed.ac.uk} \quad
 \texttt{mlap@inf.ed.ac.uk}}

\begin{document}
\maketitle

\begin{abstract}
The availability of large-scale datasets has driven the development of neural models that create summaries from single documents, for \textit{generic} purposes.
When using a summarization system, users often have \textit{specific} intents with various language realizations, which, depending on the information need, can range from a single keyword to a long narrative composed of multiple questions.
Existing summarization systems, however, often either fail to support or act robustly on this \textit{query focused} summarization task.
We introduce \mbox{\textsc{LaQSum}}, the first unified text summarization system that learns \textbf{La}tent \textbf{Q}ueries from documents for abstractive summarization with any existing query forms.
Under a deep generative framework, 
our system jointly optimizes a \textit{latent query model} and a \textit{conditional language model}, 
allowing users to plug-and-play queries of any type at test time.
Despite learning from only generic summarization data and requiring no further optimization for downstream summarization tasks, our system robustly outperforms strong comparison systems across summarization benchmarks with different query types, document settings, and target domains.
\end{abstract}

\section{Introduction}
The neural encoder-decoder framework has become increasingly popular in \emph{generic} summarization (\citealt{see2017get,gehrmann2018bottom,liu2019hierarchical}, \emph{inter alia}) thanks to the availability of large-scale datasets containing hundreds of thousands of document-summary pairs. 
\emph{Query focused} summarization (QFS; \citealt{dang2005overview}) which aims to create a short summary from one or multiple document(s) that answers a specific query, 
in comparison, 
does not have training data of this magnitude;
existing QFS corpora \cite{dang2005overview,hoa2006overview,nema2017diversity,baumel2016topic} are relatively small for training large neural architectures and have been mostly used for evaluation.
Therefore, how to leverage generic summarization data for the benefit of QFS has become a research topic of interest recently \cite{xu2020abstractive,laskar2020query}. It, however, remains a challenging research question due to the absence of queries in generic summarization data. 

Early attempts in QFS sidestep this problem by seeking distant supervision from query-relevant NLP tasks \cite{xu2020query,su2020caire,laskar2020wsl}
and exploiting existing resources, including datasets and pretrained models in question answering \cite{rajpurkar2016squad,chakraborty2020biomedbert} and paraphrase
identification \cite{mrpc}.
Since these resources can also be extremely expensive to acquire \cite{bajaj2016ms}, 
recent work proposes to induce proxy queries from generic summaries \cite{xu2020abstractive}, enabling an abstractive QFS system to be trained with only query-free resources.

In this work, we note that queries can be realized in various language forms, including but not limited to one or multiple keyword(s) \cite{baumel2016topic,wikiref}, a natural question \cite{nema2017diversity} and a composition of multiple sub-queries \cite{Dang:2006}. 
These diversified query languages inevitably lead to a mismatch between the actual queries a system has to take as inputs at test time, and the queries pre-collected or generated from distant supervision for training purposes. 
% Although marge tries to find a unified representation for summaries and queries as the middle-ground, we note that discrepancy in query formats still exists between proxy and true queries, including query length and query content. 
% In fact, it is challenging to make queries and summaries completely identical on the level of language realization. 
Moreover, the scalability of these approaches is constrained in terms of handling different query types. 
For instance, the performance of an abstractive QFS system trained on proxy queries can be sensitive to the format of target queries at test time which the proxy ones were created to mimic \cite{xu2020abstractive}.
To make a summarization system work well with every new query type, one may have to re-design the proxy algorithm, re-create the proxy queries, and re-train one or more system modules, which is computationally inefficient and sometimes practically infeasible.

In this work, we aim at building an abstractive text summarization system that is robust across \textit{observed} and \textit{latent} query settings.
Particularly, we treat generic summarization data as a special case of QFS where the query is unspecified and \textit{empty},
and assume it to be the only accessible resource for \textit{both} model training and development.
To build a generalizable summarization system, we 
% Since generic summarization data provided with empty queries is the only training resource we assume accessible,
propose to model queries as \textit{discrete latent variables} from documents, and build a latent query representation that is compatible with different query language realizations as inputs.
Specifically, we formulate an abstractive summarization task as a generative process, and decompose the objective into two components:
(1) latent query modeling (i.e., generating latent query variables from a document observation) and 
(2) conditional language modeling (i.e., generating an abstractive summary conditioned on the observed document and latent queries). 
To further enable user-specified query inputs of different formats at test time, we provide a non-parametric calibration to the latent query distribution; it can be plugged into the latent variable model without re-training, and therefore enables \textit{zero}-shot QFS.

Our contributions in this work are threefold: 
we propose the first text summarization system that unifies abstractive generic summarization and QFS. 
Particularly, no query-related resource is required for model training or development;
we provide a general, deep generative formulation for text summarization, under which we validate the effectiveness of representing queries directly from input documents in the \textit{latent} space, i.e., without resorting to pipeline-style query extraction or generation;
we provide experimental results on a wide spectrum of summarization benchmarks and show that across query types, document settings, and target domains, our system achieves better results than strong comparison systems.

\section{Related Work}
\label{sec:relwork}
\label{sec:formulation}
\begin{figure*}[t]
  \centering
  \vspace{-2em}
  \includegraphics[width=16cm]{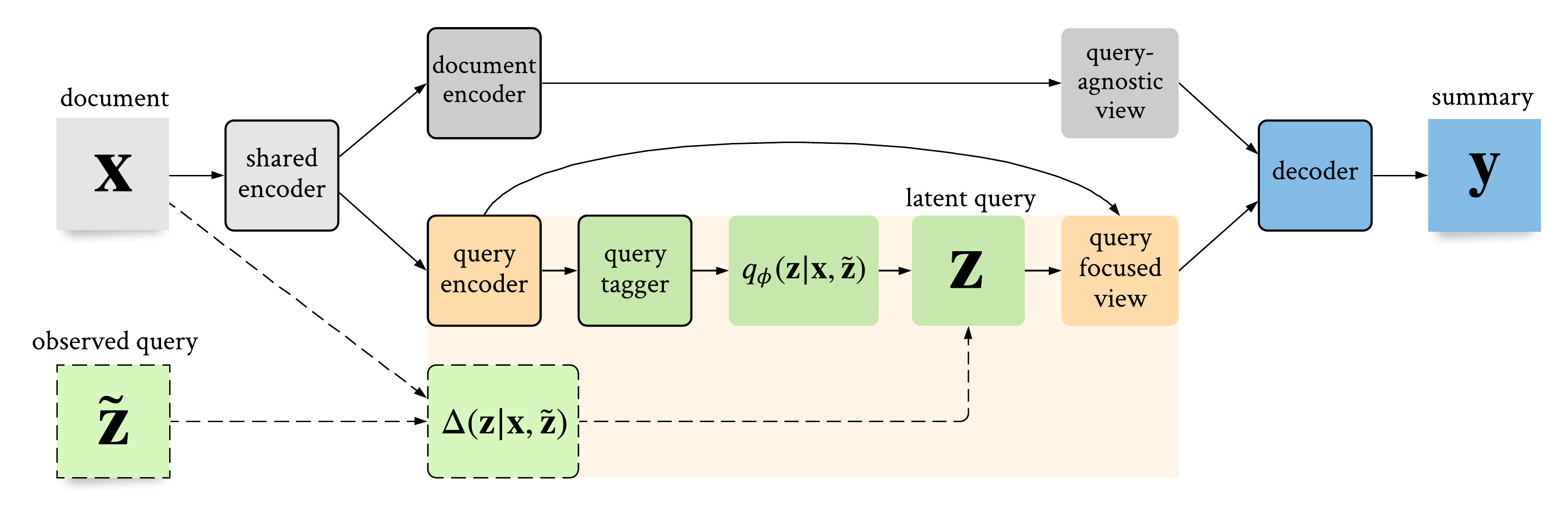}
  \caption{\label{fig:framework} Proposed text summarization framework.
Dashed lines denote non-parametric query observation modeling at testing, which is optional. 
Shadow denotes components for latent query modeling, which infers discrete latent variables as queries. 
Latent queries create a query focused view of the input document, which, together with a query-agnostic view, is input to a conditional language model for summary generation.
}
\end{figure*}
\subsection{Generic Abstractive Summarization}
\newcite{rush-2015-neural} and \newcite{nallapati-2016-abstractive} are among the first to apply the neural encoder-decoder architecture to abstractive summarization. 
\newcite{see2017get} enhance the system with a pointer-generator model where a copy mechanism is proposed to allow words in the source document to be copied directly in summary generation. 
In a bottom-up fashion,
\newcite{gehrmann2018bottom} propose to train a word-level tagging model separately; at test time, it produces content selection probabilities for each word, which are then used to restrict the copy mechanism by performing hard masking over the input document.
Inspired by \newcite{gehrmann2018bottom}, we also propose to train a tagging model for abstractive summarization. In this work, a tagger serves as a parameterized component of an inference model for latent query variables and is jointly optimized via a generic summarization task and a weakly supervised tagging task.

Another line of research in abstractive summarization propose to control text generation via topics \cite{perez2019generating,wang2020friendly}, retrieved summaries \cite{cao-2018-retrieve}, or
factual relations \cite{zhu-2021-enhancing}.
Notably, \newcite{dou2020gsum} propose a controllable system for abstractive summarization which achieves state-of-the-art performance by taking sentences extracted by another state-of-the-art \textit{extractive} system \cite{matchsum} as guidance at test time.
Despite the conceptual similarity between guidance and queries, 
we note that one fundamental difference between guided summarization and query focused summarization lies in their task objectives: guidance is created for improving generic summarization on aspects such as informativeness \cite{cao-2018-retrieve} and faithfulness \cite{zhu-2021-enhancing}, 
while QFS handles user intents of various forms in a low-resource setting 
where the lack of training data for high-quality guidance creation renders a guided system not directly applicable.

\subsection{Query Focused Summarization}
Extractive QFS, the dominant approach in QFS research, 
composes summaries by selecting \textit{central} and \textit{query-relevant} sentences in documents, based on different ways of estimating and incorporating centrality and relevance \cite{wan2007manifold,badrinath2011improving,wan2014ctsum,li2017salience,li2017cascaded}.
More recently, \newcite{xu2020query} propose a coarse-to-fine framework that leverages distant supervision from question answering for summary sentence extraction.

Abstractive QFS, compared to either its generic or extractive counterpart, has received significantly less attention,
due to generation models being particularly data-hungry
\cite{lebanoff2018adapting,liu2019hierarchical}. 
To alleviate the scarcity of QFS data, resources from a wider range of NLP tasks have been investigated for generating query focused abstracts. 
\newcite{su2020caire} rank document paragraphs against queries with a plethora of QA and machine reading datasets \cite{su2019generalizing,rajpurkar2016squad}, then summarize selected paragraphs iteratively.
Similarly, \newcite{laskar2020wsl} jointly exploit supervision from QFS data (typically reserved for evaluation) and related QA and paraphrase identification tasks.

Since external query-related resources such as QA datasets can also be costly to obtain \cite{bajaj2016ms,kwiatkowski2019natural}, 
\newcite{xu2020abstractive} assume access to only \textit{query-free} resources for abstractive QFS training.
They discover a type of connection between queries and summaries, and, alternatively, create proxy queries from generic summaries.
These proxy queries are created by selectively masking information slots in summaries. 
To minimize covariance shift, they are created to be as close to target queries as possible, on aspects such as content and length.
Despite the promising system performance and relaxed training resource assumption, 
this approach still assumes prior knowledge of the target query type at testing, and therefore relies on a development set for proxy query generation and model selection \cite{xu2020abstractive}. 
% Also, it is not sensitive to the format of user input. 
% Also, the abstractive summarization, which is based on a query model of evidence ranking and selection, 
Also, the system is particularly tailored for \textit{multi}-document QFS with an evidence selection component.

In this work, we aim at \textit{zero}-shot transfer setting for QFS tasks, i.e., we do not assume the availability of even one small QFS dataset for development purposes, as it is challenging to obtain a development set for every query language form. 
We present a system that can performing well on both single- and multi-document QFS across both observed and latent query settings.

\section{Problem Formulation}
Let \mbox{$\{(\mathcal{D}, \mathcal{Q}, \mathcal{S})\}$} denote a summarization dataset, where $\mathcal{D}$ is a document token sequence with corresponding summary $\mathcal{S}$, 
and query~$\mathcal{Q}$ additionally specifies an information request. 
We have $\mathcal{Q}=\emptyset$ in generic summarization, 
% Let \mbox{$\{(S, D)\}$} denote a
% generic summarization dataset where \mbox{$D = \{x_1, x_2,
%   \dots, x_M\}$} is an input document with corresponding summary $S$.
% In QFS, a query~$Q$ additionally specifies an information request,
% \mbox{$\{(S, D, Q)\}$}. 
% while in QFS, depending on the test set, $\mathcal{Q}$ can be in the format of a keyword (e.g., \textsl{Amnesty International}), 
% a natural question (e.g., \textsl{Is self-defense a good reason for guns ownership?}), 
% a narrative (\textsl{What is the scope of operations of Amnesty
%   International and what are the international reactions to its
%   activities?}), 
% or a combination of above. We detail these two models next.
while in QFS, depending on the test set, $\mathcal{Q}$ can be in various formats, from keywords to composite questions (see Table \ref{tab:data_stats} for examples of different query types).

In this work, we aim at building a summarization system that learns only from generic summarization data, while robustly generalizing to a range of summarization tasks at test time, including in-domain generic summarization and out-of-domain QFS tasks.
One common characteristic of generic summarization and QFS is the \textit{under-specified} user intent: even in QFS, despite $\mathcal Q \neq \emptyset$, a query usually presents incomplete guidance signals and does not fully represent a user's intent, as a query usually \textit{seeks} information, so it is unlikely for a user to specify every aspect of what is needed to compose a good summary \cite{xu2020abstractive}. 
Therefore, for both generic summarization and QFS, it is favorable that a system can identify \textit{latent} query signals from $\mathcal D$, to which, $\mathcal Q$ can, optionally, serve as an additional observation for belief update.

\paragraph{Generative Model} 
We model the observed input document $\mathcal D$ as a sequence of random variables \mbox{$\mathbf{x} = [\mathbf{x}_1; \mathbf{x}_2; \dots; \mathbf{x}_M]$} where $\mathbf{x}_i$ is a token and $M$ length of document. 
We define \textit{latent query}, a sequence of discrete latent states over the input document tokens: \mbox{$\mathbf{z} = [\mathbf{z}_1; \mathbf{z}_2; \dots; \mathbf{z}_M]$}.
Specifically, from each document token $\mathbf{x}_i$, we generate a binary query variable $\mathbf{z}_i$, whose distribution $p(\mathbf{z}_i)$ indicates the belief that $\mathbf{x}_i$ contributes to a potential query for the document $\mathcal D$. 
The output variable \mbox{$\mathbf{y} = [\mathbf{y}_1; \mathbf{y}_2; \dots; \mathbf{y}_T]$} is then generated  from $\{\mathbf{x},\mathbf{z}\}$ using teacher-forcing at training time.
Note that at test time, one may have access to additional query knowledge $\mathcal Q$; we also \textit{ground} this optional query information to the input document as discrete \textit{observed}  variables $\tilde{\mathbf{z}}=[\tilde{\mathbf{z}}_1; \tilde{\mathbf{z}}_2; \dots; \tilde{\mathbf{z}}_M]$, and 
generate $\mathbf{y}$ by additionally conditioning on $\tilde{\mathbf{z}}$ (if it exists) in an autoregessive manner. 

We consider modeling the conditional distribution  $p_\theta(\mathbf{y} \vert \mathbf{x}) $ and write down its factorization according to the generative process described above: 
\begin{align}
    p_\theta(\mathbf{y} \vert \mathbf{x})
    &=
    \sum_\mathbf{z} p_\theta(\mathbf{y}|\mathbf{z},\mathbf{x}) p_\theta(\mathbf{z}|\mathbf{x}) \\
    &=
    \sum_\mathbf{z} p_\theta(\mathbf{y}|\mathbf{z}, \mathbf{x}) \prod_i p_\theta(\mathbf{z}_i|\mathbf{x}_i)\nonumber
\end{align}

\begin{comment}
We then consider modeling the joint distribution  $p_\theta(\mathbf{x}, \mathbf{y}) $ and write down its factorization according to the generative process described above: 
\begin{align}
    p_\theta(\mathbf{x}, \mathbf{y}) 
    &= p_\theta(\mathbf{x}) p_\theta(\mathbf{y} \vert \mathbf{x})\\
    &= p_\theta(\mathbf{x}) \sum_\mathbf{z} p_\theta(\mathbf{y}|\mathbf{z},\mathbf{x}) p_\theta(\mathbf{z}|\mathbf{x})\nonumber \\
    &= p_\theta(\mathbf{x}) \sum_\mathbf{z} p_\theta(\mathbf{y}|\mathbf{z}, \mathbf{x}) \prod_i p_\theta(\mathbf{z}_i|\mathbf{x}_i)\nonumber
\end{align}
\end{comment}

\paragraph{Inference Model}  The posterior distribution of the latent variable $\mathbf{z}$ is calculated as: 
\begin{align}
    p_\theta (\mathbf{z}|\mathbf{x}, \mathbf{y}) = \frac{p_\theta(\mathbf{x},\mathbf{y},\mathbf{z})}{p_\theta(\mathbf{x},\mathbf{y})} = \frac{p_\theta(\mathbf{x},\mathbf{y},\mathbf{z})}{\sum_{\mathbf{z}} p_\theta(\mathbf{x},\mathbf{y},\mathbf{z})}.
\end{align}

However, exact inference of this posterior is computationally intractable due to the joint probability $p_\theta(\mathbf{x},\mathbf{y})$. 
We, therefore, opt for a variational posterior  $q_\phi(\mathbf{z}|\mathbf{x},\mathbf{y})$ to approximate it.
Inspired by $\beta$-VAE \cite{betavae}, we maximize the probability of generating text $\mathbf{y}$, while keeping the distance between the prior and variational posterior distributions under a small constant $\delta$: 
% Formally, we write the objective as:
\begin{align}
&\max_{\phi, \theta} \E_{(\mathbf{x}, \mathbf{y})\sim \mathcal{D}}
\left[\E_{\mathbf{z} \sim q_\phi(\mathbf{z}\vert\mathbf{x},\mathbf{y} )} \log p_\theta(\mathbf{y}\vert \mathbf{x}, \mathbf{z})
\right]\\
&\text{subject to } D_\text{KL}\left(q_\phi(\mathbf{z}\vert\mathbf{x}, \mathbf{y})\|p_\theta(\mathbf{z} \vert \mathbf{x})
\right) < \delta\label{eq:constraint}
\end{align}

Since we cannot solve Equation (\ref{eq:constraint}) directly, we apply the Karush–Kuhn–Tucker conditions (KKT; \citealt{kuhn1951nonlinear}) and cast  the constrained optimization problem into unconstrained optimization, with the following form of ELBO objective:
\begin{align}
    \mathcal{L}_{\text{ELBO}} &= 
    \E_{q_\phi(\mathbf{z}|\mathbf{x},\mathbf{y})}
    \left[\log p_\theta(\mathbf{y}|\mathbf{x},\mathbf{z})
    \right]\\
    &-\beta D_{\text{KL}}
    \left(q_\phi(\mathbf{z}|\mathbf{x},\mathbf{y}) || p_\theta(\mathbf{z}|\mathbf{x})
    \right) \nonumber
\end{align}

In this work, 
% we do not assume access to any prior query information for model learning. 
to minimize our system's dependence on testing data,
we adopt a uniform prior $p_\theta(\mathbf{z}|\mathbf{x})$, based on the hypothesis that given all instances of $\mathbf{x}$ in data (which may come from different domains), the aggregated probability of variable $\mathbf{z}$, i.e., whether a word is a query word under various contexts, follows a uniform distribution.
% i.e., the aggregated probability of a word being a query word over all contexts is $0.5$.
In this case, minimizing the KL term is equivalent to maximizing the entropy of the variational posterior.\footnote{When $p_\theta(\mathbf{z}|\mathbf{x}) \sim \mathcal{U}(a,b)$,  $D_{\text{KL}}(q_\phi(\mathbf{z}|\mathbf{x},\mathbf{y}) || p_\theta(\mathbf{z}|\mathbf{x}))
=-\mathcal{H}\left(q_\phi(\mathbf{z}|\mathbf{x})\right)+\log (b-a+1)$ always holds.}
For posterior approximation, we further assume $\mathbf{z} \indep \mathbf{y}$ and therefore $q_\phi(\mathbf{z}|\mathbf{x},\mathbf{y})=q_\phi(\mathbf{z}|\mathbf{x})$.
% and approximate $p_\theta(\mathbf{z}|\mathbf{x},\mathbf{y})$ with $q_\phi(\mathbf{z}|\mathbf{x})$. 
While this reduces the risk of exposure bias from $\mathbf{y}$ during learning, 
it is more challenging to learn meaningful latent variables as they solely condition on $\mathbf{x}$. 
We mitigate this issue by introducing a new type of weak supervision $o(\hat{\mathbf{z}}\vert \mathbf{x},\mathbf{y})$ that we are able to automatically extract from data.
This weak supervision, which is in the form of sequence tagging and will be discussed in Section \ref{sec:query_model}, leads an extra objective term for posterior regularization.
Formally, we rewrite the final optimization objective as:

% \begin{align}
%     \mathcal{L} 
%     &=
%     \underbrace{\E_{q_\phi(\mathbf{z}|\mathbf{x})}\left(\log p_\theta(\mathbf{y}|\mathbf{x},\mathbf{z})\right)
%     - \beta \mathcal{H}(q_\phi(\mathbf{z}|\mathbf{x})) \nonumber}_{\text{ELBO}} \\
%     &+ \omega \underbrace{\mathcal{H}(\hat{\mathbf{z}}, \mathbf{z})}_{\text{Weak Supervision}}
% \end{align}
\begin{align}
    \mathcal{L} 
    % &= \mathcal{L}_{\text{ELBO}} + \omega \mathcal{L}_{\text{weak}}\\
    &=
    \underbrace{\E_{q_\phi(\mathbf{z}|\mathbf{x})}
    \left[\log p_\theta(\mathbf{y}|\mathbf{x},\mathbf{z})
    \right]}_{\text{conditional language modeling}} \label{eq:objective}\\
    &+
    \underbrace{\beta \mathcal{H}\left(
    q_\phi(\mathbf{z}|\mathbf{x})
    \right) 
    - 
    \omega \mathcal{H}\left(
    o(\hat{\mathbf{z}} \vert \mathbf{x}, \mathbf{y}), q_\phi(\mathbf{z} \vert \mathbf{x})
    \right)}_{\text{latent query modeling}}\nonumber
\end{align}
where $\mathcal{H}(\cdot)$ denotes posterior entropy and $\mathcal{H}(\cdot,\cdot)$ denotes cross entropy for posterior regularization from weak supervision.
Particularly, we decompose a summarization task into two modeling objectives, 
\textit{latent query modeling} and \textit{conditional language modeling}.
Inside query modeling, hyperparameter $\omega$ controls the influence from the weak supervision $\hat{\mathbf{z}}$, while $\beta$ controls the strength of a form of label smoothing (see Section \ref{sec:query_model} for details). 

% \begin{comment}
\begin{table*}[t]
\centering
\footnotesize
\setlength{\tabcolsep}{3pt}
\begin{tabular}{lll}
  \thickhline
Error & Data & Example \\
\thickhline
\multirow{3}{*}{Type I} & Summary & 
\makecell[l]{Real Madrid slump to \textbf{defeat} against Athletic Bilbao. \\Solitary goal from Aritz Aduriz enough to give the Basques victory. \\Bayern Munich continue Bundesliga domination.}\\
~ & Document & ..., with a convincing 4-1 win over Lens at the Parc \textbf{de} Princes. ... \\
~ & Annotation & ..., with\#0 a\#0 convincing\#0 4-1\#0 win\#0 over\#0 Lens\#0 at\#0 the\#0 Parc\#0 \textbf{de\#1} Princes.\#0 ...\\
\hline
\multirow{6}{*}{Type II} & Summary & 
\makecell[l]{A man in suburban \textbf{Boston} is selling snow online to customers in warmer states. 
\\For \$89, he will ship 6 pounds of snow in an insulated Styrofoam box.}\\
~ & Document & 
\makecell[l]{... For \$89, self-styled entrepreneur Kyle Waring will ship you 6 pounds of \textbf{Boston-area} snow 
\\in an insulated Styrofoam box ...} \\
~ & Annotation & 
\makecell[l]{... For\#1 \$89,\#1 self-styled\#0 entrepreneur\#0 Kyle\#0 Waring\#0 will\#1 ship\#1 you\#0 6\#1 pounds\#1 \\
of\#1 \textbf{Boston-area\#0} snow\#1 in\#1 an\#1 insulated\#1 Styrofoam\#1 box\#1 ...} \\
\thickhline
\end{tabular}
% \vspace*{-.2cm}
\caption{\label{tab:lcs} 
Error analysis for longest common sub-sequence tagging scheme (LCS; \citealt{gehrmann2018bottom}) with instances from CNN/DM validation set. 
Document tokens and binary annotations are separated with \#. 
Bold fonts denote erroneous annotations and their corresponding token in summaries/documents.
}
\end{table*}
\paragraph{Neural Parameterization}
We show our framework in Figure \ref{fig:framework}.
We parameterize the two modeling objectives in Equation (\ref{eq:objective}) with a \textit{latent query model} and an \textit{conditional language model}, respectively.
The query model is optimized to estimate  latent query $\mathbf z$ from its input variable $\mathbf x$.
At inference time, it, optionally, conditions on prior query knowledge $\hat{\mathbf{z}}$ (when available in QFS data).
% Prior query knowledge represents what we know of a query without access to  $\mathcal D$ or $\mathcal S$.
% For instance, in QFS, a query is given in each sample and the tokens in it can serve as prior knowledge.
% As prior query knowledge is not provided in generic summarization, we assume $\mathbf{z} \indep Q$.
% or synthesize $Q$ from $\mathcal{D}$ with heuristics.
The conditional language model, on the other hand, augments the standard encoder-decoder structure to take as input two encoding sources: a \textit{query-agnostic view} and a \textit{query focused view}.
We call them two \textit{views} as they are separate encodings of the same input $\mathcal D$, with the only difference lies in the direct dependence on $\mathbf{z}$ generated from the query model.
In contrast to using only the query-focused view, 
maintaining two views allows the original document context to be retained as a complementary feature space.
Finally, a decoder is used to generate summary $\mathbf{y}$ auto-regressively.
Different from previous abstractive QFS formulation \cite{xu2020abstractive}, we jointly train these two models in a fully differentiable summarization system.

\section{Latent Query Model}
\label{sec:query_model}
In this section we detail the neural parameterization for latent query modeling in Equation (\ref{eq:objective}),
including the architecture for query inference and document representation in a query focused view. 
We also introduce learning from weak supervision $o(\hat{\mathbf{z}} \vert \mathbf{x}, \mathbf{y})$ and testing with belief update.

\paragraph{Inference Network for Latent Queries} 
% Query prior does not consider either the local document context, or its corresponding summary.
We construct a neural network model to infer query belief for each token in the input document. 
% by calculating its posterior $q(z_i=1 | Q, D, S)$.
% In this work we propose to use query embeddings to inject prior query knowledge:
% \begin{equation}
    % q_i = E_q * b_i.
% \end{equation}
% where $E_q \in \R^{d_h \times 2}$ is an embedding matrix optimized during training. As the input to the query belief updater, we append $q_i$ to the word representation $h_i$: $h'_i = [h_i, q_i]$.
Given a contextual token representation matrix \mbox{$\mathbf{H}_q \in \R^{M \times d_h}$}, 
we project it to $\R^{M \times 2}$ with a two-layer MLP as a scoring function:
% we project it to a low-dimensional space $\R^{M \times 2}$ with a two-layer MLP as scoring function:
% \begin{gather}
%     \mathbf{H}_s =
%     \relu (\mathbf{W}_h \mathbf{H}_q^{\intercal}+\mathbf{b}_h) \\
%     \bm{\pi}=\mathbf{W}_s \mathbf{H}_s^{\intercal} + \mathbf{b}_s \label{eq:pi}
% \end{gather}
\begin{gather}
    \mathbf{H}_s =
    \relu ( \mathbf{H}_q \mathbf{W}_h+\mathbf{b}_h^{\intercal}) \\
    \bm{\pi}= \mathbf{H}_s \mathbf{W}_s + \mathbf{b}_s^{\intercal} \label{eq:pi}
\end{gather}
where $\mathbf{W}_h \in \R^{d_h \times d_h}$, $\mathbf{b}_h\in\R^{d_h\times 1}$, $\mathbf{W}_s \in \R^{d_h \times 2}$ and $\mathbf{b}_s\in\R^{2\times 1}$ are learnable model parameters.

Let $G(0)$ denote the standard Gumbel distribution, and $g_{\ell} \sim G(0), \ell \in [0,1]$ are i.i.d. gumbel noise. We normalize $\bm{\pi}$ to form a variational distribution as:
% \begin{equation}
% \scalebox{0.91}{
%   $ q_\phi(\mathbf{z}|\mathbf{x}) =\softmax([\bm{\pi}_0 + g_0,  \bm{\pi}_1+g_1])$.
% }
% \end{equation}
% \begin{equation}
%   q_\phi(\mathbf{z}|\mathbf{x}) =\softmax([\bm{\pi}_0 + g_0,  \bm{\pi}_1+g_1]).
% \end{equation}
\begin{align}
    q_\phi(\mathbf{z}_i=
    \ell|\mathbf{x}) &=\text{softmax}_{\ell}([\bm{\pi}_0 + g_0, \bm{\pi}_1+g_1])
    \nonumber
    \\
    % &=\frac{\exp((\bm{\pi}_0 + g_0)/\tau)}{\exp((\bm{\pi}_0 + g_0)/\tau)+\exp((\bm{\pi}_1 + g_1)/\tau)}
    &=\frac{\exp((\bm{\pi}_{\ell} + g_{\ell})/\tau)}{\sum_{{\ell'} \in [0,1]}\exp((\bm{\pi}_{\ell'} + g_{\ell'})/\tau)}
\end{align}
where $\tau$ is the temperature controlling how close $q_\phi(\mathbf{z}|\mathbf{x})$ is  to $\argmax_{\ell} q_\phi(\mathbf{z}|\mathbf{x})$, and is optimized on the development set.
Note that gumbel noise is only applied during learning and is set to its mode, i.e., 0, for inference. 
% Since Transformer can be seen as a case of a graph neural network applied on a  fully-connected word graph \cite{dwivedi2020generalization}, our query belief networks can be viewed as a message passing process that propagates prior query labels on a graph for query expansion \cite{de1998query}.
\paragraph{Query Focused View}
In addition to the query-agnostic document representation, a direct encoding of the input  document $\mathcal D$ adopted by a plethora of summarization models (which we leave to Section \ref{sec:abs}), we further introduce a query focused view,
% We call it a \textit{view} as it is a separate encoding of the same input $\mathcal D$ in addition to the original document encoding (which we call \textit{document view} and will be introduced in Section \ref{sec:abs}). 
an encoding factorized via latent queries $\mathbf z$.
% \begin{equation}
%     \mathbf{Q} = f_\phi(\mathcal{D}) = f_\phi(\mathbf{H}_q, \mathbf{z}).
% \end{equation}

Specifically, for the $i$th token, we take the continuous relaxation of its discrete latent variable $\mathbf{z}_i$, and ground the query to the input document in the representation space via:\footnote{We also experimented with drawing hard samples of $\mathbf{z}$ via 
% (1) non-differentiable $\argmax$ and
the straight-through trick \cite{sst} which is differentiable with biased gradient estimation. However, it did not yield better results than continuous relaxation.}
\begin{equation}
    % \mathbf{Z} = \mathbf{H}_{q} * \I(B_{;, 1}>\delta)
    % \mathbf{Q}_i = \mathbf{H}_{q, i} \cdot q_\phi(\mathbf{z}_i=1 \vert \mathbf{x}).
    \mathbf{Q}_i = q_\phi(\mathbf{z}_i=1 \vert \mathbf{x}) \cdot  \mathbf{H}_{q, i}.
\end{equation}
% Here, we propose to do continuous relaxation 
% where $\I$ is an indicator function, i.e., 1 when the input condition is true and 0 otherwise. Threshold $\delta$ is a hyperparameter controling the confidence level required to be a query token. We optimize $\delta$ on the development set.
As we can see, the query focused view explicitly models the dependency on latent queries.
% From the learning perspective, this factorization  provides an extra gradient direction $\frac{\partial \textbf{Q}}{\partial q_{\phi}}$ that can be more efficient than $\frac{\partial \textbf{Q}}{\partial \mathcal{H}}$, since it acts as 
From the learning perspective, this factorization leads to the following partial derivatives of the query focused view with respect to the query encoder states:
% \begin{align}
% \frac{\partial \mathbf{Q}}{\partial \mathbf{H}_q} 
% &=q_\phi^{(0)} \cdot (\mathbf{H}_q \cdot q_\phi^{(1)}) + q_\phi^{(1)} \cdot \mathbf{J} \\
% &=q_\phi^{(0)} \cdot \mathbf{Q} + q_\phi^{(1)} \cdot \mathbf{J},
% \end{align}
\begin{comment}
\begin{align}
\frac{\partial \mathbf{Q}_i}{\partial \mathbf{H}_{q,i}} 
&=
\left(1-q_\phi^{(1)}\right) \cdot 
\frac{\partial \Delta \pi} {\partial \mathbf{H}_{q,i}} \cdot
% \left(
\mathbf{H}_q \cdot q_\phi^{(1)}
% \right) 
+ q_\phi^{(1)} \cdot \mathbf{1} 
\nonumber
\\
&=
\left(1-q_\phi^{(1)}\right)
\cdot 
% \left(
\frac{\partial \Delta \pi}{\partial \mathbf{H}_{q,i}} 
\odot \mathbf{Q}_i 
% \right)
+ q_\phi^{(1)} \cdot \mathbf{1}
\end{align}
\end{comment}
\begin{equation}
\frac{\partial \mathbf{Q}_i}{\partial \mathbf{H}_{q,i}} 
=
 \underbrace{
 \left(1-q_\phi^{(1)}\right)
 }_{\text{carry gate}}
 \cdot 
 \frac{\partial \Delta \pi} {\partial \mathbf{H}_{q,i}} 
\odot
\mathbf{Q}_i
+ 
\underbrace{q_\phi^{(1)}}_{\mathclap{\text{transform gate}}}
\cdot 
\mathbf{1} 
% \nonumber
\end{equation}
where $q_\phi^{(\ell)}$ is the shorthand for the variational probability of $\mathbf{z}_{i}=\ell \vert \mathbf{x}$, and $\Delta \pi=\bm{\pi}_1-\bm{\pi}_0$ (see Equation (\ref{eq:pi})).
$\mathbf{1}$ denotes an all-one vector. 
This can be seen as a special case of highway networks \cite{highway_networks} with a zero-map transform function, where the transform gate $q_\phi^{(1)}$ controls the information compression rate.

% From the learning perspective, this factorization  provides an extra gradient direction $\frac{\partial \textbf{Q}}{\partial q_{\phi}}$:
% that can be more efficient than $\frac{\partial \textbf{Q}}{\partial \mathcal{H}}$, since it acts as 

\paragraph{Sequence Tagging as Weak Supervision}
\begin{table*}[t]
\centering
\footnotesize
\setlength{\tabcolsep}{3pt}
\begin{tabular}{llllllp{6.7cm}}
  \thickhline
Dataset & Task & Domain &
Size &
D/Q/S Tokens &
Query Type &
Query Example \\
\thickhline
CNN/DM & SDS & News & 11,490 & 760.5/0.0/45.7 & Empty & $\emptyset$ \\
WikiRef & SDS & Wiki & 12,000 & 398.7/6.7/36.2
& Keywords & \textit{Marina Beach, Incidents} \\
Debatepedia & SDS & Debate & 1,000 & 
66.4/10.0/11.3 & Question & \textit{Is euthanasia better than withdrawing life support?}
\\
% \hline
\multirow{2}{*}{DUC 2006} &
\multirow{2}{*}{MDS} &
\multirow{2}{*}{Cross} & 
\multirow{2}{*}{1,250 (50)} & 
\multirow{2}{*}{699.3/32.8/250} &
\multirow{4}{*}{Composite} & 
\multirow{4}{6.7cm}{
\hspace{-1ex}
\textbf{\textcolor{gred}{\textit{Amnesty International}} - \textcolor{gblue}{\textit{What is the scope of operations of Amnesty International and what are the international reactions to its activities?}}}
}
\\
\\
\multirow{2}{*}{DUC 2007} & \multirow{2}{*}{MDS} & \multirow{2}{*}{Cross} &
\multirow{2}{*}{1,125 (45)} & 
\multirow{2}{*}{540.3/30.5/250}
\\
\\
TD-QFS & MDS & Medical & 7,099 (50) & 182.9/3.0/250& Title & \textit{Alzheimer’s Disease}
\\
\thickhline
\end{tabular}
\vspace*{-.2cm}
\caption{\label{tab:data_stats} Test data statistics.
SDS and MDS stand for single- and multi-document summarization, respectively. 
Size refers to number of documents for single-document test set; for MDS, we additionally specify number of clusters in brackets.
In the composite query example, red and blue fonts denote its title and narrative, respectively.
}
\end{table*}

% Title: \textit{Amnesty International} \\
% Narrative: \
% \textit{What is the scope of operations of Amnesty International}\\\textit{and what are the international reactions to its activities?}
Since our system is fully differentiable, it is possible to optimize latent queries solely based on conditional language modeling.
In this work, we additionally propose to exploit sequence tagging as weak supervision. 
This can be advantageous since it imposes extra regularization via posterior constraints to prevent its collapse, in which case the decoder may learn to ignore the query focused view and instead solely rely on the query agnostic view.
% by making all-one predictions.

A challenge with applying sequence tagging to summarization is the absence of gold query annotation in training data. Inspired by \newcite{gehrmann2018bottom}, we align summary and document by searching for their longest common sub-sequences (LCS). 
Nevertheless, we note that there exist a few drawbacks hindering its direct application as weak supervision.
Primarily, LCS treats a document as a word sequence and annotates it at word-level, while our tagging model (built on top of a pretrained language model) operates on subwords.
Apart from this incompatibility of granularity, LCS can lead to false-negative word labels (Type II error), and it is also sensitive to false-positive cases (Type I error) since a summary is seen as a \textit{character} sequence in LCS.
We provide examples of these two error types in Table \ref{tab:lcs}.
% For instance, when a summary contains a number  \textsl{68} and a document contains \textsl{8 teams} in its leading sentences, the document word \textsl{8} will be annotated as positive since the summary token sequence \textsl{68} contains \textsl{8}, 
% despite the lack of contextual relevance between these two numbers. 
We propose a simple but effective solution to fix the abovementioned issues: 
we first byte-pair encode (BPE; \citealt{bpe}) documents and summaries, and then search LCS over paired document-summary BPE sequences.
Compared to the other way around, doing BPE as the first step allows finer-grained unit matching (which reduces Type II error), while still retains basic semantics (which reduces Type I error).
We annotate BPEs in the extracted LCS as 1 and the rest as 0. 
Note that if there exist multiple identical LCS, only the one appearing at the earliest document position is tagged as positive. 
We refer to this query tagging scheme as \textsc{Bpe-Lcs}.
% \yumo{Question: this is doubtful in our setting: shouldn't we label every LCS? Why only the earliest one?}

\paragraph{Training}
We use a cross entropy loss for sequence tagging, 
% where 1~denotes that a token is a query token (and 0~otherwise).
% According to Equation (\ref{eq:objective}), we also add an entropy term for the variational posterior:
with a posterior entropy term in Equation (\ref{eq:objective}):
\begin{align}
    \mathcal{L}_{\text{query}} 
    &=-\omega \mathcal{L}_{\text{tag}}+\beta \mathcal{L}_{\text{entropy}}
    \label{eq:query_loss}
    \\
	&=
	-\sum_{j=1}^N\sum_{i=1}^M 
	\bigl(
	\bigl(\omega \hat{\mathbf{z}}^j_i-\beta q_\phi^{(1)} \bigr)
	\log q_\phi^{(1)} \nonumber
	\\
	&+ 
	\bigl(
	\omega \bigl(1 - \hat{\mathbf{z}}^{j}_i\bigr)-\beta q_\phi^{(0)}\bigr)
	\log q_\phi^{(0)}
	\bigr) \nonumber
\end{align}
where $\hat{\mathbf{z}}_i$ is a binary annotation automatically assigned via \mbox{\textsc{Bpe-Lcs}$(\mathcal{D}, \mathcal{S})$}. 
% $q_\phi^{(\ell)}$ is the shorthand for the variational probability of $\mathbf{z}^{j}_{i}=\ell  \vert \mathbf{x}$.
As we can see, the entropy term smooths the weak annotation $\hat{\mathbf{z}}_i$, with a dynamic smoothing strength dependent on $q_\phi$.
We optimize $\omega, \beta$ on the development set.

We notice that at the initial training phase, the under-optimized tagger produces inaccurate posterior $q_\phi(\mathbf{z}_i \vert \mathbf{x})$, and, consequently, hurts learning an abstractive summarization model which heavily relies on a high-quality query focused view. 
To tackle this issue, we propose a \textit{posterior dropout} mechanism: with a probability $\delta$, we replace the estimated posterior with the weak supervision $o(\hat{\mathbf{z}} \vert \mathbf{x})$. 
We initialize $\delta$ to 1.0, i.e., only $o(\hat{\mathbf{z}} \vert \mathbf{x})$ is used,
and the tagger is supervised via Equation (\ref{eq:query_loss}). 
We then linearly anneal $\delta$ over optimization steps to $\delta_{\text{end}}$, 
so the gradients from the summarization objective (which will be introduced in Section \ref{sec:abs}) can further optimize the tagger jointly.

\begin{comment}
\begin{align}
    \mathcal{L}_{\text{query}} 
    &=-\omega \mathcal{L}_{\text{tag}}+\beta \mathcal{L}_{\text{entropy}}\\
	&=
	\sum_{j=1}^N\sum_{i=1}^M 
	\bigl(
	\hat{\mathbf{z}}_i \log q_\phi(\mathbf{z}^{j}_{i}=1 \vert \mathbf{x}) \\
	&+ 
	[\beta q_\phi(\mathbf{z}^{j}_{i}=0 \vert \mathbf{x}) -\omega(1 - \hat{\mathbf{z}}^{j}_i)]
	\log q_\phi(\mathbf{z}^{j}_{i}=0 \vert \mathbf{x})
	\bigr) \nonumber
\end{align}
\end{comment}

\begin{table}[h]
\footnotesize
\tabcolsep=0.1cm
\bgroup
\centerline{
% \begin{small}
\begin{tabular}{lccc}
\hline
% \multirow{2}*{\textbf{Systems}} 
% & \multicolumn{3}{c}{\textbf{CNN/DM}} \\
\textbf{Systems} & {R-1} & {R-2} & {R-L} \\
\thickhline
{\sc Oracle}
&55.8 &33.2 &51.8 
\\
{\sc Lead}
&40.4	&17.6	&36.7
\\
\hline
% \multicolumn{4}{c}{\textit{Extractive}}
\textit{Extractive}
\\
\hline
% \hdashline
{\sc BertExt} \cite{liu2019text}
&43.9 &20.3 &39.9
\\
{\sc MatchSum} \cite{matchsum}
&43.9	&20.6	&39.8
\\
\hline
% \multicolumn{4}{c}{\textit{Abstractive}}
\textit{Abstractive}
\\
\hline
% \hdashline
{\sc PTGen} \cite{see2017get}
&39.5 &17.3 &36.4
\\
{\sc BottomUp} \cite{gehrmann2018bottom}
&41.2 &18.7 &38.4
\\
{\sc BertAbs} \cite{liu2019text}
&41.7 &19.4 &38.8
\\
{\sc Bart} \cite{bart}
&44.2   &21.3   &40.9
\\
{\sc GSum} \cite{dou2020gsum}
&45.9	&22.3	&42.5
\\
{\sc GSum} (our implementation)
& 45.0	& 21.9	& 41.8
\\
\hline
{\sc  LaQSum}
& \textbf{45.1}	& \textbf{22.0}	& \textbf{41.9}
\\
\thickhline
\end{tabular}
}
\egroup
\caption{\label{tab:cnndm} 
  Supervised performance on \textbf{CNN/DM} test set.
  R-1, R-2 and R-L stand for the F1 score of ROUGE~1, 2, and L, respectively. \textsc{GSum} (our implementation) uses the same training configurations as our model.
}
\end{table}
\begin{table}[h]
\footnotesize
\tabcolsep=0.1cm
\bgroup
\centerline{
% \begin{small}
\begin{tabular}{lccc}
\hline
% \multirow{2}*{\textbf{Systems}} 
% & \multicolumn{3}{c}{\textbf{WikiRef}}\\
% ~ & {R-1} & {R-2} & {R-L} \\
\textbf{Systems} & {R-1} & {R-2} & {R-L} \\
\thickhline
{\sc Oracle}
&54.5 &37.5 &48.5
\\
{\sc Lead}
&26.3 &10.5 &21.8
\\
{\sc LexRank}
&29.9	&12.3	&26.1
\\
\hline
% \multicolumn{4}{c}{\textit{Supervised (Extractive)}}
\textit{Supervised (Extractive)}
\\
\hline
% \hdashline
{\sc Transformer} \cite{wikiref}
&28.1   &12.8   &23.8
\\
{\sc BertExt} \cite{wikiref}
&36.0 &18.8 &30.7
\\
\hline
% \multicolumn{4}{c}{\textit{Zero-shot Abstractive}}
\textit{Zero-shot Abstractive}
\\
\hline
% \hdashline
{\sc Bart} \cite{bart}
&30.0	&12.2	&26.0
% \\
% {\sc GSum+Query}
% &30.2	&12.5	&26.3
\\
{\sc GSum+Query$_\mathcal E$}
% &30.3	&\bf{12.5}	&26.4
&30.2	&{12.5}	&26.3
\\
\hline
{\sc  LaQSum}
&\bf{31.1}	&\bf{12.6}	&\bf{27.1}
\\
\thickhline
\end{tabular}
}
\egroup
\caption{\label{tab:wikiref} 
  Zero-shot performance on \textbf{WikiRef} test set (with keywords as queries).
  R1, R2 and RL stand for the F1 score of ROUGE~1, 2, and L, respectively. 
%   \textsc{GSum} (ours) uses the guidance sentences selected by LexRank.
}
\end{table}
\begin{table}[h]
\footnotesize
\tabcolsep=0.05cm
\bgroup
\centerline{
% \begin{small}
\begin{tabular}{lccc}
\hline
\textbf{Systems} & {R-1} & {R-2} & {R-L} \\
\thickhline
% {\sc Oracle}
% \\
{\sc Lead}
&18.1	&\hspace{0.5em}5.6	&15.9
\\
{\sc LexRank}
& 17.4	&\hspace{0.5em}5.3	&15.1\\
\hline
\textit{Supervised (Abstractive)}
\\
\hline
{\sc Dda} \cite{laskar2020query} & \hspace{0.5em}7.4	&\hspace{0.5em}2.8	&\hspace{0.5em}7.2 \\
{\sc BertAbs+Rank} \cite{qg_on_debatepedia}	& 19.2	&10.6	&17.9 \\
{\sc BertAbs+Concat} \cite{laskar2020query} & 26.4 &11.9 &25.1 \\
\hline
\textit{Zero-shot Abstractive}
\\
\hline
{\sc BertAbs}$^\dagger$ \cite{liu2019text} &	13.3	&\hspace{0.5em}2.8	&\hspace{0.5em}2.8
\\
{\sc Bart} \cite{bart}
&21.4	&\hspace{0.5em}6.3	&18.4
\\
% {\sc GSum+Query}
% &23.2	&\hspace{0.5em}\bf{7.4}	&20.0
% \\
% {\sc GSum+GRSum}
{\sc GSum+Query$_\mathcal E$}
&21.2	&\hspace{0.5em}6.2	&18.2
\\
\hline
{\sc  LaQSum}
&\bf{23.5}	&\hspace{0.5em}\bf{7.2}    &\bf{20.6}
\\
\thickhline
\end{tabular}
}
\egroup
\caption{\label{tab:debatepedia} 
  Zero-shot performance on \textbf{Debatepedia} test set (with natural questions as queries).
  R1, R2 and RL stand for the F1 score of ROUGE~1, 2, and L, respectively. 
%   \textsc{GSum} (ours) uses the guidance sentences selected by LexRank.
  $\dagger$ denotes models optimized on XSum \cite{xsum} and numbers are borrowed from \newcite{laskar2020query}.
}
\end{table}

\paragraph{Testing}
To plug-and-play query focus during testing, 
we model the optional query knowledge one may have access to, $\mathcal Q$, with an query belief updator $\Delta(\mathbf{z}_i|\mathbf{x}, \tilde{\mathbf{z}})$.
% During testing, 
Specifically, when no prior query knowledge $\mathcal Q$ is accessible (i.e., in generic summarization), 
one may assume \textit{zero} increment for all tokens' query belief.
% : \mbox{$\Delta p(\mathbf{z}_i=1|Q) = 0, i \in [1, M]$}. 
While in QFS, we have prior knowledge $\mathcal Q \neq \emptyset$ that some tokens come with high query belief and therefore should be biased over:
% In this work, we model this prior knowledge with a distribution $p'(\mathbf{z}_i|Q)$,  
we set $\Delta(\mathbf{z}_i=1|\mathbf{x}, \tilde{\mathbf{z}})=1.0, \forall w_i \in \textsc{Bpe-Lcs}(\mathcal{D}, \mathcal{Q})$, and the rest to zero.
% To incorporate prior query information, we interpolate it with the variational posterior: 
% \begin{equation}
%     q_\theta(\mathbf{z}_i|\mathbf{x}, Q) = \lambda * p'(\mathbf{z}_i|Q) * (1-\lambda) q_\theta(\mathbf{z}_i|\mathbf{x})
% \end{equation}

We further incorporate prior query information via a simple calibration as:
\begin{align}
    q_\phi(\mathbf{z}_i=1\vert \mathbf{x}, \tilde{\mathbf{z}}) 
    &= \max \{1.0, 
    \\
    &q_\phi(\mathbf{z}_i=1\vert \mathbf{x})+
    \Delta(\mathbf{z}_i=1|\mathbf{x}, \tilde{\mathbf{z}})\}. \nonumber
    % \\
    % &q_\phi(\mathbf{z}_i=0\vert \mathbf{x}, Q) = 1.0 - q_\phi(\mathbf{z}_i=1\vert \mathbf{x}, Q) \nonumber
\end{align}
\begin{comment}
\begin{equation}
\hspace*{-.7cm}
q_\phi(\mathbf{z}_i\vert \mathbf{x}, Q)
\hspace*{-.1cm}=\hspace*{-.1cm} \\
\begin{cases} 
    1 & 
    \hspace*{-.15cm}p'(\mathbf{z}_i=1\vert Q)=1\\
	 q_\phi(\mathbf{z}_i\vert \mathbf{x}) &
	 \hspace*{-.15cm}\text{otherwise.}
\end{cases}
\end{equation}
\end{comment}
% An alternative is to linearly interpolating the variational posterior \mbox{$ q_\phi(\mathbf{z}\vert \mathbf{x})$} 
% and the prior \mbox{$p'(\mathbf{z}|Q)$} with a coefficient as hyper-parameter \cite{xu2020query}. 
Note that our adopt a \textit{non-parametric} query belief calibration, as we do not assume the availability of a development set of each query type for hyper-parameter optimization. This enables \textit{zero}-shot transfer to QFS tasks with different settings.

\section{Conditional Language Model}
\label{sec:abs}
In this section we introduce conditional language modeling which models the expectation of the log-likelihood of a summary word sequence over the variational posterior distribution in Equation (\ref{eq:objective}).
As shown in Figure \ref{fig:framework}, we adopt an encoder-decoder architecture tailored for text summarization with latent queries.

\paragraph{Encoder}
We encoder the same input document into two different views, a query-agnostic view, and a query focused view. 
Therefore, our encoder module consists of three encoders: a shared encoder, a document encoder, and a query encoder. 
The intuition is straightforward: 
since both views are created from the same document, we use a shared encoder for general document understanding which also reduces model parameters.
The shared document representation is then input to the other two separate encoders to encode high-level view-specific features. 
Each encoder contains one or multiple Transformer layers \cite{vaswani2017attention}, which is composed of a multi-head attention (MHA) layer and a feed-forward (FFN) layer:
\begin{align}
\text{\small
$\HE$}
&\text{\small
$=\layernorm\bigl(\HE+\mha\bigl(\HE,\HE,\HE\bigr)\bigr)$}
\nonumber\\
\text{\small
$\HE$}
&
\text{\small
$=\layernorm\bigl(\HE+\ffn\bigl(\HE\bigr)\bigr)$}.
\end{align}
where $\layernorm$ denotes layer normalization. 
As outputs of the encoding module, 
the query focused view $\mathbf{Q}$ directly conditions on latent query variables, while the query-agnostic view $\mathbf{D}$ retains original contexts.

\begin{table*}[t]
\footnotesize
\tabcolsep=0.15cm
\bgroup
\def\arraystretch{1.0}
\centerline{
\begin{tabular}{lccc ccc ccc}
\hline
\multirow{1}*{\textbf{Models}} 
& \multicolumn{3}{c}{\textbf{DUC 2006}}
& \multicolumn{3}{c}{\textbf{DUC 2007}}
& \multicolumn{3}{c}{\textbf{TD-QFS}}\\
~ & {R-1} & {R-2} & {R-SU4} 
& {R-1} & {R-2} & {R-SU4} 
& {R-1} & {R-2} & {R-SU4}
\\
\hline
{\sc  Gold} 
& 45.7 &11.2 &  17.0 
& 47.9 &14.1 &  19.1  
& --- & --- & --- \\
{\sc  Oracle} 
& 40.6 &\hspace*{1ex}9.1 &  14.8 
& 41.8 &10.4 & 16.0 
& 44.9 &18.9 &23.0\\
{\sc  Lead} 
& 32.1 &\hspace*{1ex}5.3 & 10.4 
& 33.4 & \hspace*{1ex}6.5 & 11.3 
& 33.5 &\hspace*{1ex}5.2 &10.4\\
% {\sc TermFreq}
% &36.5	&\hspace*{1ex}7.0	&12.6
% &38.5	&\hspace*{1ex}9.0	&14.2
% &35.7	&\hspace*{1ex}6.5	&12.0\\
{\sc  LexRank}
&34.2 &\hspace*{1ex}6.4 & 11.4 
&35.8 &\hspace*{1ex}7.7 & 12.7 
&35.3 &\hspace*{1ex}7.6 &12.2\\
\hline
\textit{Distantly Supervised}
\\
\hline
% \hdashline
{\sc QuerySum}$^\ast$ \cite{xu2020query}
& 41.6 & \hspace*{1ex}9.5 & 15.3
& 43.3 & 11.6 & 16.8
& 44.3 & 16.1 & 20.7
\\
% \hline
% \textit{Abstractive}
% \\
% \hline
{\sc Bart-Caq} \cite{su2020caire} 
&38.3 &\hspace*{1ex}7.7 &12.9
&40.5 &\hspace*{1ex}9.2 &14.4
& --- & --- & --- \\
{\sc PQSum} \cite{laskar2020wsl}
&{40.9} &\hspace*{1ex}9.4 & 14.8 
& 42.2 & 10.8 & 16.0
& --- & --- & --- \\
\hline
\textit{Few- or Zero-shot Abstractive}
\\
\hline
% \hdashline
{\sc MargeSum}$^\dagger$
\cite{xu2020abstractive}
&{40.2}     &\hspace*{.8ex}9.7	&{15.1}	
&{42.5}	&12.0   &16.9
&{45.5}	&{16.6}	&{20.9}
\\
% \hline
% \textit{Zero-shot Abstractive}
% \\
% \hline
{\textsc{Bart}} \cite{bart}
&38.3	&\hspace*{1ex}7.8	&13.1	
&40.2	&\hspace*{1ex}9.9	&14.6
&45.1	&16.9	&21.4
\\
% {\sc GSum+Query}
% &37.7	&\hspace*{1ex}7.9	&12.9	
% &39.3	&\hspace*{1ex}9.7	&14.2
% &43.8	&14.8	&19.3
% \\
{\sc GSum+Query$_\mathcal E$}
&38.1	&\hspace*{1ex}7.9	&13.1	
&39.5	&\hspace*{1ex}9.5	&14.3
&45.5	&18.0	&\textbf{22.4}
\\
{\textsc{LaQSum}}
&\textbf{39.1}	&\hspace*{1ex}\textbf{8.5}	&\textbf{13.7}
&\textbf{40.4}	&\textbf{10.2}	&\textbf{15.0}
% &\textbf{45.8}	&\textbf{17.4}	&\textbf{21.8}
% &45.7	&17.4	&21.7
&\textbf{45.7}	&\textbf{18.1}	&{22.1}
\\
\thickhline
\end{tabular}
}
\egroup
\caption{\label{tab:mds}
 Zero-shot performance on multi-document QFS test sets \textbf{DUC} (with composed queries) and \textbf{TD-QFS} (with titles as queries).
 $\ast/\dagger$: extractive/few-shot system.
%  $\dagger$: developed on QFS data.
 R1, R2 and R-SU4 stand for the F1 score of ROUGE~1, 2, and SU4, respectively. 
}
\end{table*}

\paragraph{Decoder} We adopt a similar decoder structure as in \newcite{dou2020gsum} to handle multiple inputs. Instead of incorporating pre-extracted guidance in \newcite{dou2020gsum}, our decoder attends to the two encoded views of the same document sequentially:
\begin{align}
\text{\small$\HD$} & \text{\small{$=\layernorm\bigl(\HD+\mha\bigl(\HD,\HD,\HD\bigr)\bigr)$}}
\nonumber\\
\text{\small$\HD$} & \text{\small{$=\layernorm\bigl(\HD+\mha\bigl(\HD,\mathbf{Q},\mathbf{Q}\bigr)\bigr)$}}
\nonumber\\
\text{\small$\HD$} & \text{\small{$=\layernorm\bigl(\HD+\mha\bigl(\HD,\mathbf{D},\mathbf{D}\bigr)\bigr)$}}
\nonumber\\
\text{\small$\HD$} & \text{\small{$=\layernorm\bigl(\HD+\ffn\bigl(\HD\bigr)\bigr)$}}.
\end{align}
After taking in the context of the previous generation, the decoder first fuses in query signals from $\mathbf{Q}$, which then drives the incorporation of original document context $\mathbf{D}$. 
The final summary generation objective is calculated auto-regressively as:
\begin{equation}
    \mathcal{L}_{\text{lm}} = 
    \sum_{j=1}^N
    \sum_{t=1}^{T} \log
    p_\theta \left(\mathbf{y}_{t}| \mathbf{y}_{<t}, \mathbf{D}, \mathbf{Q} \right)
    \label{eq:xe_seq_loss}
\end{equation}
which is jointly trained with the query model (see Equation (\ref{eq:query_loss})) as: $\mathcal{L}=\mathcal{L}_{\text{lm}}+\mathcal{L}_{\text{query}}$.

% Therefore, the final objective is:
% \begin{equation}
% \mathcal{L} = \mathcal{L}_{\text{gen}} + \omega * \mathcal{L}_{\text{tag}}
% \end{equation}
% We optimize the tagging loss coefficient $\omega$ on the development set. 

\section{Experimental Setup}
\paragraph{Datasets}
We used CNN/DM \cite{hermann2015teaching}, a generic single-document summarization dataset containing news articles and associated highlights, for model training and development (with 287,227/13,368 instances).
For model evaluation, we evaluated our system on CNN/DM test set (11,490 instances) under a supervised setting. 
We also performed experiments on QFS under a \textit{zero}-shot transfer setting, on five test sets with various formats of queries, domains, and document settings, including WikiRef \cite{wikiref}, Debatepedia \cite{nema2017diversity}, DUC 2006-07, and TD-QFS \cite{baumel2016topic}.
Statistics for all test sets are given in Table \ref{tab:data_stats}.
 Note that we do not assume any development data in QFS, which departs from \newcite{xu2020abstractive}.

% Transfer learning for QFS includes \textit{zero}-shot and \textit{few}-shot experiments.
\paragraph{Implementation Details}
The shared encoder consists of 11 Transformer layers.
The document and query encoder has one separate Transformer layer each.
The shared encoder, document encoder, and decoder are initialized with a pretrained \textsc{Bart} model \cite{bart}, while the query encoder is randomly initialized.
We used 4 GeForce RTX 2080 GPUs for training; we set the batch size to 8 (i.e., one sample for each GPU), and accumulate gradients every 32 steps.
Following the standard configurations for \textsc{Bart} finetuning on CNN/DM, we used a learning
rate of $3\times 10^{-5}$ for 20,000 optimization steps, with a warmup-step of 500.
Due to memory constraints, we used half float precision for efficient training and also set the maximum length of an input document to 640 tokens, with the excess clipped. 
We set $\beta=0.1$ and $\omega=10$ in the learning objective. 
We used $\tau=0.9$ for latent query modeling. For the proposed posterior dropout, we annealed the dropout rate $\delta$ from 1.0 to 0.5 over the whole training session.

% \input{tab_ablation}
%% All data for Ablation Study
\begin{table*}[t]
\footnotesize
\tabcolsep=0.1cm
\bgroup
\def\arraystretch{1.0}
\centerline{
% \begin{small}
% \begin{tabular}{@{}l@{~}@{~}c@{~~}c@{~~}cc@{~~}c@{~~}c@{}}  % The first row
\begin{tabular}{lc c c c c c}
\hline
\multirow{1}*{\textbf{Models}} 
& {\textbf{CNN/DM}}
& {\textbf{WikiRef}}
& {\textbf{Debatepedia}}
& {\textbf{DUC 2006}}
& {\textbf{DUC 2007}}
& {\textbf{TD-QFS}}
\\
\hline
{\textsc{LaQSum}}
&{41.9}
&{27.1}
&{20.6}
&{13.7}
&{15.0}
&{22.1}
\\
{\hspace*{2ex}-$\Delta(\hat{\mathbf{z}}|\mathbf{x},\mathbf{z})$}
& --- 
&$\downarrow$0.2
&$\downarrow$0.6
&$\downarrow$0.6	
&$\downarrow$1.3
&$\downarrow$0.4
\\
{\hspace*{2ex}-Joint training}
&$\downarrow$0.4	
&$\downarrow$2.8
&$\downarrow$2.8	
&$\downarrow$1.6	
&$\downarrow$1.7
&$\downarrow$0.4
\\
{\hspace*{2ex}-Weak supervision}
&$\downarrow$0.7	
&$\downarrow$0.5	
&$\downarrow$1.3
&$\downarrow$0.2	
&$\downarrow$0.3
&$\downarrow$0.0
\\
{\hspace*{2ex}-Dual view}
&$\downarrow$2.5
&$\hspace{0.5em}\downarrow$10.5	
&$\downarrow$6.6	
&$\downarrow$1.8	
&$\downarrow$2.5	
&$\downarrow$2.8
\\
{\hspace*{2ex}-Posterior dropout}
&$\downarrow$0.8	
&$\downarrow$0.7
&$\downarrow$1.2	
&$\downarrow$0.2	
&$\downarrow$0.5
&$\uparrow$0.1
\\
\hline
\end{tabular}
% \end{small}
}
\egroup
\caption{\label{tab:ablation} 
  Ablation results for our abstractive summarization system.
  $\uparrow/\downarrow$:
  absolute performance increase/decrease in ROUGE-L (on CNN/DM, WikiRef and Debatepedia) or ROUGE-SU4 (on DUC 2006-07 and TD-QFS).
}
\end{table*}
\section{Results}
\paragraph{Generic Summarization}
Table \ref{tab:cnndm} summarizes our results on CNN/DM. 
The first block in the table includes an \textsc{Oracle} extractive system as an upper bound. \textsc{Lead} baseline take the first 3 sentences in a document as a summary. 

The second block presents two extractive systems. \textsc{BertExt} \cite{liu2019text} is the first system using as a pretrained encoder \cite{devlin2019bert} for text summarization. \textsc{MatchSum} is a state-of-the-art extractive system extracting an optimal set of sentences via text matching.

The third block includes various abstractive systems (see Section \ref{sec:relwork} for an overview). Particularly, \textsc{PTGen} \cite{see2017get} and \textsc{BottomUp} \cite{gehrmann2018bottom} do not use pretrained LMs, 
while \textsc{BertAbs} is built on a pretrained \textsc{Bert} encoder,
and \textsc{GSum} \cite{dou2020gsum}, similar to our work, is initialized with pretrained \textsc{Bart} parameters \cite{bart}. 
Our system outperforms standard the \textsc{Bart} finetuned on CNN/DM by a fairly large margin, which demonstrates the effectiveness of modeling the query focused view with latent queries even for generic summarization.
Under the same training resources and configurations,
it also performs on par with \textsc{GSum}, a state-of-the-art abstractive model, despite being significantly more computationally efficient, as no access to high-quality guidance sentences (which are produced by another well-trained extractive system, e.g., \textsc{MatchSum}) is required.

\paragraph{Single-Document QFS}
We show results on WikiRef and Debatepedia in Table \ref{tab:wikiref}
 and Table \ref{tab:debatepedia}, respectively, for single-document QFS evaluation.

 In the first block of these two tables, we show two \textit{unsupervised} extractive baselines: \textsc{Lead} and \textsc{LexRank} which estimates sentence-level centrality via Markov Random Walk on graphs. 
 The second block presents various \textit{supervised} systems on WikiRef and Debatepedia.
 Note that no abstractive QFS system has been evaluated on WikiRef, while Debatepedia is a short-document, short-summary dataset mainly for abstractive summarization. 
 
The third block of the two tables highlights system performance in the \textit{zero}-shot transfer setting, including \textsc{Bart} and \textsc{GSum}.
Particularly, \textsc{GSum} requires guidance from a \textit{generic} extractive summarization system, 
which is hard to obtain due to the data scarcity in QFS. 
Also, it is not straightforward how a given query can be incorporated to generate query-specific summaries from \textsc{GSum}. 
% we build two variants of \textsc{GSum} for testing. 
% To make it query oriented, we build \textsc{GSum+Query}, which directly takes the given query as guidance. 
% Since \textsc{GSum} is not directly applicable to QFS, 
To adapt it to QFS test settings, we build \textsc{GSum+Query$_{\mathcal E}$}, 
where we employ an unsupervised query focused extractive system to pre-extract the top$K$ ranked sentences for each testing document as its guidance.
Specifically, we choose a query focused version of \textsc{LexRank} described in \newcite{xu2020query}, which is well-performed on extractive QFS tasks by jointly estimating sentence centrality and query relevance \cite{wan2008using}.
% We set $K=3$ for WikiRef (which has longer documents) and $K=1$ for Debatepedia (which has shorter documents). 

 In an end-to-end fashion, our system achieves the highest ROUGE scores on both datasets in the zero-shot transfer setting. 
 Compared to the results on generic data, our system shows a clearer edge over systems without latent query modeling.
 
\paragraph{Multi-Document QFS} 
To apply a summarization system trained on \textit{single}-document data to a \textit{multi}-document setting, we adopt a simple iterative generation approach \cite{baumel2018query}: 
we first rank documents in a cluster via query term frequency, and then generate summaries iteratively for each document.
The final summary for the whole cluster is composed by concatenating document-level summaries.\footnote{
An alternative is to generate a long summary at once. However, this requires a model to be trained on a MDS dataset, or at least a proxy one \cite{xu2020abstractive}. 
Since we build our system also for single-document summarization, we choose to generate and then compose.
}
Repetitive generated sentences are skipped to remove redundancy.

Table \ref{tab:mds} presents results on multi-document QFS datasets. 
The first block reports performance of two upper-bound systems, \textsc{Gold} and \textsc{Oracle}, and two unsupervised systems taken from \newcite{xu2020abstractive}.
The second block contains previous \textit{distantly supervised} approaches.
\textsc{QuerySum}
\cite{xu2020query} is state-of-the-art extractive system which adopts QA datasets for a coarse-to-fine salience estimation process.
On the abstractive side, \textsc{Bart-Caq} \cite{su2020caire} uses an
ensembled QA model for answer evidence extraction, and then use finetuned
\textsc{Bart} \cite{bart} to iteratively generate summaries from paragraphs.  
\textsc{PQSum} \cite{laskar2020wsl} uses finetuned \textsc{BertSum} to generate summaries for each document in a cluster,
and a QA model for summary sentenc re-ranking.

The third block compares our model with a start-of-the-art \textit{few}-short approach,
\textsc{MargeSum} \cite{xu2020abstractive} which requires a small QFS development set, and \textit{zero}-short systems including \textsc{Bart} and \textsc{GSum+Query}$_\mathcal E$.
As we can see, without recourse to expensive QA/QFS annotations, our system, achieves significantly better results than \textsc{Bart-Caq} which exploits QA data as external training resources, on DUC test sets (except in ROUGE-1 on DUC 2007); on TD-QFS, it surpasses \textsc{MargeSum} which uses QFS data for proxy query generation and model development, across all metrics. 
Also, our system outperforms strong zero-shot abstractive systems including \textsc{Bart} and \textsc{GSum+Query}$_\mathcal E$ on all three datasets.

\paragraph{Ablation Studies}
We provide the results of ablation studies on \textsc{LaqSum} in Table \ref{tab:ablation}.
Removing query belief update at test time (-$\Delta(\hat{\mathbf{z}}|\mathbf{x},\mathbf{z})$) hurts model performance on QFS test sets, demonstrating the usefulness of incorporating query information via simple calibration on the variational posterior distribution. 
When it comes to learning meaningful latent queries that benefit summarization tasks,
relying on only tagging (-Joint training, where we adopt $\argmax$ to stop gradients from the generation loss), 
or generation (-Weak supervision, where we set $\omega=0$) significantly decreases performance. 
We conclude that latent query learning performs a trade-off between exploiting \textit{direct but weak} supervision from the tagging objective (i.e., based on synthetic token annotation), 
and exploring the \textit{natural but indirect} supervision from the generation objective (i.e., based on human-written summaries). 
Removing the query agnostic view (-Dual view) causes significant performance drop as it keeps the original document context that the decoder can possibly leverage, especially when the query model is not well-performed. 
This is also supported by solely using the \textit{estimated} posterior to create query focused view for training (-Posterior dropout), which also hurts model performance as it leads to more severe error propagation to the downstream generation model.

\section{Conclusion}
In this work we provide a deep generative formulation for text summarization, 
and present a general text summarization system that supports generating both generic and query focused abstracts.
Under this formulation, queries are represented as discrete latent variables, whose approximated posterior distribution can be, optionally, calibrated with additional query observations during testing without further adaptation.
As a result, our system does not rely on any query-related resource. 
Experimental results across datasets of various characteristics show that the proposed system yields strong performance on generic summarization, and state-of-the-art performance on zero-shot abstractive QFS.

% \nobibliography{custom}
\bibliography{custom.bib}

\begin{thebibliography}{48}
\expandafter\ifx\csname natexlab\endcsname\relax\def\natexlab#1{#1}\fi

\bibitem[{Abdullah and Chali(2020)}]{qg_on_debatepedia}
Deen~Mohammad Abdullah and Yllias Chali. 2020.
\newblock Towards generating query to perform query focused abstractive
  summarization using pre-trained model.
\newblock In \emph{Proceedings of the 13th International Conference on Natural
  Language Generation}, pages 80--85, Dublin, Ireland.

\bibitem[{Badrinath et~al.(2011)Badrinath, Venkatasubramaniyan, and
  Veni~Madhavan}]{badrinath2011improving}
Rama Badrinath, Suresh Venkatasubramaniyan, and CE~Veni~Madhavan. 2011.
\newblock Improving query focused summarization using look-ahead strategy.
\newblock In \emph{Proceedings of the 33rd European Conference on Advances in
  Information Retrieval}, pages 641--652, Dublin, Ireland.

\bibitem[{Bajaj et~al.(2016)Bajaj, Campos, Craswell, Deng, Gao, Liu, Majumder,
  McNamara, Mitra, Nguyen et~al.}]{bajaj2016ms}
Payal Bajaj, Daniel Campos, Nick Craswell, Li~Deng, Jianfeng Gao, Xiaodong Liu,
  Rangan Majumder, Andrew McNamara, Bhaskar Mitra, Tri Nguyen, et~al. 2016.
\newblock {MS MARCO}: A human generated machine reading comprehension dataset.
\newblock \emph{arXiv preprint arXiv:1611.09268}.

\bibitem[{Baumel et~al.(2016)Baumel, Cohen, and Elhadad}]{baumel2016topic}
Tal Baumel, Raphael Cohen, and Michael Elhadad. 2016.
\newblock Topic concentration in query focused summarization datasets.
\newblock In \emph{Proceedings of the 30th AAAI Conference on Artificial
  Intelligence}, pages 2573--2579, Phoenix, Arizona.

\bibitem[{Baumel et~al.(2018)Baumel, Eyal, and Elhadad}]{baumel2018query}
Tal Baumel, Matan Eyal, and Michael Elhadad. 2018.
\newblock Query focused abstractive summarization: Incorporating query
  relevance, multi-document coverage, and summary length constraints into
  seq2seq models.
\newblock \emph{arXiv preprint arXiv:1801.07704}.

\bibitem[{Cao et~al.(2018)Cao, Li, Li, and Wei}]{cao-2018-retrieve}
Ziqiang Cao, Wenjie Li, Sujian Li, and Furu Wei. 2018.
\newblock Retrieve, rerank and rewrite: Soft template based neural
  summarization.
\newblock In \emph{Proceedings of the 56th Annual Meeting of the Association
  for Computational Linguistics (Volume 1: Long Papers)}, pages 152--161,
  Melbourne, Australia.

\bibitem[{Chakraborty et~al.(2020)Chakraborty, Bisong, Bhatt, Wagner, Elliott,
  and Mosconi}]{chakraborty2020biomedbert}
Souradip Chakraborty, Ekaba Bisong, Shweta Bhatt, Thomas Wagner, Riley Elliott,
  and Francesco Mosconi. 2020.
\newblock {BioMedBERT}: A pre-trained biomedical language model for qa and ir.
\newblock In \emph{Proceedings of the 28th International Conference on
  Computational Linguistics}, pages 669--679, Online.

\bibitem[{Dang(2005)}]{dang2005overview}
Hoa~Trang Dang. 2005.
\newblock Overview of duc 2005.
\newblock In \emph{Proceedings of the 2005 Document Understanding Conference},
  pages 1--12, Vancouver, Canada.

\bibitem[{Dang(2006)}]{Dang:2006}
Hoa~Trang Dang. 2006.
\newblock {DUC 2005}: Evaluation of question-focused summarization systems.
\newblock In \emph{Proceedings of the Workshop on Task-Focused Summarization
  and Question Answering}, pages 48--55, Stroudsburg, PA, USA.

\bibitem[{Devlin et~al.(2019)Devlin, Chang, Lee, and
  Toutanova}]{devlin2019bert}
Jacob Devlin, Ming-Wei Chang, Kenton Lee, and Kristina Toutanova. 2019.
\newblock {BERT}: Pre-training of deep bidirectional transformers for language
  understanding.
\newblock In \emph{Proceedings of the 2019 Conference of the North American
  Chapter of the Association for Computational Linguistics: Human Language
  Technologies}, pages 4171--4186, Minneapolis, Minnesota.

\bibitem[{Dolan and Brockett(2005)}]{mrpc}
William~B Dolan and Chris Brockett. 2005.
\newblock Automatically constructing a corpus of sentential paraphrases.
\newblock In \emph{Proceedings of the Third International Workshop on
  Paraphrasing}, pages 9--16, Jeju Island, Korea.

\bibitem[{Dou et~al.(2020)Dou, Liu, Hayashi, Jiang, and Neubig}]{dou2020gsum}
Zi-Yi Dou, Pengfei Liu, Hiroaki Hayashi, Zhengbao Jiang, and Graham Neubig.
  2020.
\newblock {GSum}: A general framework for guided neural abstractive
  summarization.
\newblock \emph{arXiv preprint arXiv:2010.08014}.

\bibitem[{Gehrmann et~al.(2018)Gehrmann, Deng, and Rush}]{gehrmann2018bottom}
Sebastian Gehrmann, Yuntian Deng, and Alexander Rush. 2018.
\newblock Bottom-up abstractive summarization.
\newblock In \emph{Proceedings of the 2018 Conference on Empirical Methods in
  Natural Language Processing}, pages 4098--4109, Brussels, Belgium.

\bibitem[{Hermann et~al.(2015)Hermann, Ko\v{c}isk\'{y}, Grefenstette, Espeholt,
  Kay, Suleyman, and Blunsom}]{hermann2015teaching}
Karl~Moritz Hermann, Tom\'{a}\v{s} Ko\v{c}isk\'{y}, Edward Grefenstette, Lasse
  Espeholt, Will Kay, Mustafa Suleyman, and Phil Blunsom. 2015.
\newblock Teaching machines to read and comprehend.
\newblock In \emph{Proceedings of the 28th International Conference on Neural
  Information Processing Systems}, page 1693–1701, Cambridge, MA, USA.

\bibitem[{Higgins et~al.(2017)Higgins, Matthey, Pal, Burgess, Glorot,
  Botvinick, Mohamed, and Lerchner}]{betavae}
I.~Higgins, Lo{\"i}c Matthey, A.~Pal, Christopher~P. Burgess, Xavier Glorot,
  M.~Botvinick, S.~Mohamed, and Alexander Lerchner. 2017.
\newblock beta-vae: Learning basic visual concepts with a constrained
  variational framework.
\newblock In \emph{ICLR}.

\bibitem[{Hoa(2006)}]{hoa2006overview}
TD~Hoa. 2006.
\newblock Overview of duc 2006.
\newblock In \emph{Proceedings of the 2006 Document Understanding Conference},
  New York, USA.

\bibitem[{Jang et~al.(2016)Jang, Gu, and Poole}]{sst}
Eric Jang, Shixiang Gu, and Ben Poole. 2016.
\newblock Categorical reparameterization with gumbel-softmax.
\newblock \emph{arXiv preprint arXiv:1611.01144}.

\bibitem[{Kuhn et~al.(1951)Kuhn, Tucker et~al.}]{kuhn1951nonlinear}
HW~Kuhn, AW~Tucker, et~al. 1951.
\newblock Nonlinear programming.
\newblock In \emph{Proceedings of the Second Berkeley Symposium on Mathematical
  Statistics and Probability}.

\bibitem[{Kwiatkowski et~al.(2019)Kwiatkowski, Palomaki, Redfield, Collins,
  Parikh, Alberti, Epstein, Polosukhin, Devlin, Lee
  et~al.}]{kwiatkowski2019natural}
Tom Kwiatkowski, Jennimaria Palomaki, Olivia Redfield, Michael Collins, Ankur
  Parikh, Chris Alberti, Danielle Epstein, Illia Polosukhin, Jacob Devlin,
  Kenton Lee, et~al. 2019.
\newblock Natural questions: a benchmark for question answering research.
\newblock \emph{Transactions of the Association for Computational Linguistics},
  7:453--466.

\bibitem[{Laskar et~al.(2020{\natexlab{a}})Laskar, Hoque, and
  Huang}]{laskar2020query}
Md~Tahmid~Rahman Laskar, Enamul Hoque, and Jimmy Huang. 2020{\natexlab{a}}.
\newblock Query focused abstractive summarization via incorporating query
  relevance and transfer learning with transformer models.
\newblock In \emph{Canadian Conference on Artificial Intelligence}, pages
  342--348. Springer.

\bibitem[{Laskar et~al.(2020{\natexlab{b}})Laskar, Hoque, and
  Huang}]{laskar2020wsl}
Md~Tahmid~Rahman Laskar, Enamul Hoque, and Jimmy~Xiangji Huang.
  2020{\natexlab{b}}.
\newblock {WSL}-{DS}: Weakly supervised learning with distant supervision for
  query focused multi-document abstractive summarization.
\newblock In \emph{Proceedings of the 28th International Conference on
  Computational Linguistics}, pages 5647--5654, Online.

\bibitem[{Lebanoff et~al.(2018)Lebanoff, Song, and Liu}]{lebanoff2018adapting}
Logan Lebanoff, Kaiqiang Song, and Fei Liu. 2018.
\newblock Adapting the neural encoder-decoder framework from single to
  multi-document summarization.
\newblock In \emph{Proceedings of the 2018 Conference on Empirical Methods in
  Natural Language Processing}, pages 4131--4141, Brussels, Belgium.

\bibitem[{Lewis et~al.(2020)Lewis, Liu, Goyal, Ghazvininejad, Mohamed, Levy,
  Stoyanov, and Zettlemoyer}]{bart}
Mike Lewis, Yinhan Liu, Naman Goyal, Marjan Ghazvininejad, Abdelrahman Mohamed,
  Omer Levy, Veselin Stoyanov, and Luke Zettlemoyer. 2020.
\newblock {BART}: Denoising sequence-to-sequence pre-training for natural
  language generation, translation, and comprehension.
\newblock In \emph{Proceedings of the 58th Annual Meeting of the Association
  for Computational Linguistics}, pages 7871--7880, Online.

\bibitem[{Li et~al.(2017{\natexlab{a}})Li, Lam, Bing, Guo, and
  Li}]{li2017cascaded}
Piji Li, Wai Lam, Lidong Bing, Weiwei Guo, and Hang Li. 2017{\natexlab{a}}.
\newblock Cascaded attention based unsupervised information distillation for
  compressive summarization.
\newblock In \emph{Proceedings of the 2017 Conference on Empirical Methods in
  Natural Language Processing}, pages 2081--2090, Brussells, Belgium.

\bibitem[{Li et~al.(2017{\natexlab{b}})Li, Wang, Lam, Ren, and
  Bing}]{li2017salience}
Piji Li, Zihao Wang, Wai Lam, Zhaochun Ren, and Lidong Bing.
  2017{\natexlab{b}}.
\newblock Salience estimation via variational auto-encoders for multi-document
  summarization.
\newblock In \emph{Proceedings of the 31th AAAI Conference on Artificial
  Intelligence}, pages 3497--3503, San Francisco, California, USA.

\bibitem[{Liu and Lapata(2019{\natexlab{a}})}]{liu2019hierarchical}
Yang Liu and Mirella Lapata. 2019{\natexlab{a}}.
\newblock Hierarchical transformers for multi-document summarization.
\newblock In \emph{Proceedings of the 57th Annual Meeting of the Association
  for Computational Linguistics}, pages 5070--5081, Florence, Italy.

\bibitem[{Liu and Lapata(2019{\natexlab{b}})}]{liu2019text}
Yang Liu and Mirella Lapata. 2019{\natexlab{b}}.
\newblock Text summarization with pretrained encoders.
\newblock In \emph{Proceedings of the 2019 Conference on Empirical Methods in
  Natural Language Processing and the 9th International Joint Conference on
  Natural Language Processing}, Hong Kong, China.

\bibitem[{Nallapati et~al.(2016)Nallapati, Zhou, dos Santos, Gulcehre, and
  Xiang}]{nallapati-2016-abstractive}
Ramesh Nallapati, Bowen Zhou, Cicero dos Santos, Caglar Gulcehre, and Bing
  Xiang. 2016.
\newblock Abstractive text summarization using sequence-to-sequence {RNN}s and
  beyond.
\newblock In \emph{Proceedings of The 20th {SIGNLL} Conference on Computational
  Natural Language Learning}, pages 280--290, Berlin, Germany.

\bibitem[{Narayan et~al.(2018)Narayan, Cohen, and Lapata}]{xsum}
Shashi Narayan, Shay~B. Cohen, and Mirella Lapata. 2018.
\newblock Don{'}t give me the details, just the summary! topic-aware
  convolutional neural networks for extreme summarization.
\newblock In \emph{Proceedings of the 2018 Conference on Empirical Methods in
  Natural Language Processing}, pages 1797--1807, Brussels, Belgium.

\bibitem[{Nema et~al.(2017)Nema, Khapra, Laha, and
  Ravindran}]{nema2017diversity}
Preksha Nema, Mitesh~M. Khapra, Anirban Laha, and Balaraman Ravindran. 2017.
\newblock Diversity driven attention model for query-based abstractive
  summarization.
\newblock In \emph{Proceedings of the 55th Annual Meeting of the Association
  for Computational Linguistics}, pages 1063--1072, Vancouver, Canada.

\bibitem[{Perez-Beltrachini et~al.(2019)Perez-Beltrachini, Liu, and
  Lapata}]{perez2019generating}
Laura Perez-Beltrachini, Yang Liu, and Mirella Lapata. 2019.
\newblock Generating summaries with topic templates and structured
  convolutional decoders.
\newblock In \emph{Proceedings of the 57th Annual Meeting of the Association
  for Computational Linguistics}, pages 5107--5116, Florence, Italy.

\bibitem[{Rajpurkar et~al.(2016)Rajpurkar, Zhang, Lopyrev, and
  Liang}]{rajpurkar2016squad}
Pranav Rajpurkar, Jian Zhang, Konstantin Lopyrev, and Percy Liang. 2016.
\newblock {SQuAD}: 100,000+ questions for machine comprehension of text.
\newblock In \emph{Proceedings of the 2016 Conference on Empirical Methods in
  Natural Language Processing}, pages 2383--2392, Sydney, Australia.

\bibitem[{Rush et~al.(2015)Rush, Chopra, and Weston}]{rush-2015-neural}
Alexander~M. Rush, Sumit Chopra, and Jason Weston. 2015.
\newblock A neural attention model for abstractive sentence summarization.
\newblock In \emph{Proceedings of the 2015 Conference on Empirical Methods in
  Natural Language Processing}, pages 379--389, Lisbon, Portugal.

\bibitem[{See et~al.(2017)See, Liu, and Manning}]{see2017get}
Abigail See, Peter~J. Liu, and Christopher~D. Manning. 2017.
\newblock Get to the point: Summarization with pointer-generator networks.
\newblock In \emph{Proceedings of the 55th Annual Meeting of the Association
  for Computational Linguistics}, pages 1073--1083, Vancouver, Canada.

\bibitem[{Sennrich et~al.(2016)Sennrich, Haddow, and Birch}]{bpe}
Rico Sennrich, Barry Haddow, and Alexandra Birch. 2016.
\newblock Neural machine translation of rare words with subword units.
\newblock In \emph{Proceedings of the 54th Annual Meeting of the Association
  for Computational Linguistics (Volume 1: Long Papers)}, pages 1715--1725,
  Berlin, Germany.

\bibitem[{Srivastava et~al.(2015)Srivastava, Greff, and
  Schmidhuber}]{highway_networks}
Rupesh~Kumar Srivastava, Klaus Greff, and J{\"u}rgen Schmidhuber. 2015.
\newblock Training very deep networks.
\newblock In \emph{Proceedings of the 28th International Conference on Neural
  Information Processing Systems-Volume 2}, pages 2377--2385, Montreal, Quebec,
  Canada.

\bibitem[{Su et~al.(2019)Su, Xu, Winata, Xu, Kim, Liu, and
  Fung}]{su2019generalizing}
Dan Su, Yan Xu, Genta~Indra Winata, Peng Xu, Hyeondey Kim, Zihan Liu, and
  Pascale Fung. 2019.
\newblock Generalizing question answering system with pre-trained language
  model fine-tuning.
\newblock In \emph{Proceedings of the 2nd Workshop on Machine Reading for
  Question Answering}, pages 203--211, Hong Kong, China.

\bibitem[{Su et~al.(2020)Su, Xu, Yu, Siddique, Barezi, and Fung}]{su2020caire}
Dan Su, Yan Xu, Tiezheng Yu, Farhad~Bin Siddique, Elham Barezi, and Pascale
  Fung. 2020.
\newblock {CA}i{RE}-{COVID}: A question answering and query-focused
  multi-document summarization system for {COVID}-19 scholarly information
  management.
\newblock In \emph{Proceedings of the 1st Workshop on {NLP} for {COVID}-19 at
  {EMNLP} 2020}, Online.

\bibitem[{Vaswani et~al.(2017)Vaswani, Shazeer, Parmar, Uszkoreit, Jones,
  Gomez, Kaiser, and Polosukhin}]{vaswani2017attention}
Ashish Vaswani, Noam Shazeer, Niki Parmar, Jakob Uszkoreit, Llion Jones,
  Aidan~N Gomez, {\L}ukasz Kaiser, and Illia Polosukhin. 2017.
\newblock Attention is all you need.
\newblock In \emph{Advances in Neural Information Processing Systems}, pages
  6000--6010.

\bibitem[{Wan(2008)}]{wan2008using}
Xiaojun Wan. 2008.
\newblock Using only cross-document relationships for both generic and
  topic-focused multi-document summarizations.
\newblock \emph{Information Retrieval}, 11(1):25--49.

\bibitem[{Wan et~al.(2007)Wan, Yang, and Xiao}]{wan2007manifold}
Xiaojun Wan, Jianwu Yang, and Jianguo Xiao. 2007.
\newblock Manifold-ranking based topic-focused multi-document summarization.
\newblock In \emph{Proceedings of the 20th International Joint Conference on
  Artificial Intelligence}, pages 2903--2908, Hyderabad, India.

\bibitem[{Wan and Zhang(2014)}]{wan2014ctsum}
Xiaojun Wan and Jianmin Zhang. 2014.
\newblock {CTSUM}: extracting more certain summaries for news articles.
\newblock In \emph{Proceedings of the 37th international ACM SIGIR Conference
  on Research \& Development in Information Retrieval}, pages 787--796, New
  York, United States.

\bibitem[{Wang et~al.(2020)Wang, Duan, Zhang, Wang, Tian, Chen, and
  Zhou}]{wang2020friendly}
Zhengjue Wang, Zhibin Duan, Hao Zhang, Chaojie Wang, Long Tian, Bo~Chen, and
  Mingyuan Zhou. 2020.
\newblock Friendly topic assistant for transformer based abstractive
  summarization.
\newblock In \emph{Proceedings of the 2020 Conference on Empirical Methods in
  Natural Language Processing}, pages 485--497, Online.

\bibitem[{Xu and Lapata(2020{\natexlab{a}})}]{xu2020abstractive}
Yumo Xu and Mirella Lapata. 2020{\natexlab{a}}.
\newblock Abstractive query focused summarization with query-free resources.
\newblock \emph{arXiv preprint arXiv:2012.14774}.

\bibitem[{Xu and Lapata(2020{\natexlab{b}})}]{xu2020query}
Yumo Xu and Mirella Lapata. 2020{\natexlab{b}}.
\newblock Coarse-to-fine query focused multi-document summarization.
\newblock In \emph{Proceedings of the 2020 Conference on Empirical Methods in
  Natural Language Processing}, pages 3632--3645, Online.

\bibitem[{Zhong et~al.(2020)Zhong, Liu, Chen, Wang, Qiu, and Huang}]{matchsum}
Ming Zhong, Pengfei Liu, Yiran Chen, Danqing Wang, Xipeng Qiu, and Xuanjing
  Huang. 2020.
\newblock Extractive summarization as text matching.
\newblock In \emph{Proceedings of the 58th Annual Meeting of the Association
  for Computational Linguistics}, pages 6197--6208, Online.

\bibitem[{Zhu et~al.(2021)Zhu, Hinthorn, Xu, Zeng, Zeng, Huang, and
  Jiang}]{zhu-2021-enhancing}
Chenguang Zhu, William Hinthorn, Ruochen Xu, Qingkai Zeng, Michael Zeng,
  Xuedong Huang, and Meng Jiang. 2021.
\newblock Enhancing factual consistency of abstractive summarization.
\newblock In \emph{Proceedings of the 2021 Conference of the North American
  Chapter of the Association for Computational Linguistics: Human Language
  Technologies}, pages 718--733, Online.

\bibitem[{Zhu et~al.(2019)Zhu, Dong, Wei, Qin, and Liu}]{wikiref}
Haichao Zhu, Li~Dong, Furu Wei, Bing Qin, and Ting Liu. 2019.
\newblock Transforming wikipedia into augmented data for query-focused
  summarization.
\newblock \emph{arXiv preprint arXiv:1911.03324}.

\end{thebibliography}
\bibliographystyle{acl_natbib}
\end{document}